\documentclass[runningheads]{llncs}
\usepackage{graphicx}
\usepackage{amsmath}
\usepackage{amssymb}

\usepackage{float}
\usepackage{orcidlink}

\begin{document}
\title{Detecting and clustering swallow events in esophageal long-term high-resolution manometry}

\titlerunning{Automated detection and clustering of swallows in LTHRM}
%
\author{Alexander Geiger\inst{1}\orcidlink{0009-0003-6910-5820}\and
Lars Wagner\inst{1}\orcidlink{0000-0002-3021-4152} \and
Daniel Rueckert\inst{2,4}\orcidlink{0000-0002-5683-5889} \and
Dirk Wilhelm\inst{1,3}\orcidlink{0000-0002-2972-9802} \and
Alissa Jell\inst{1,3}\orcidlink{0000-0002-7153-3803}}
\authorrunning{A. Geiger et al.}
%

\institute{Technical University of Munich, TUM School of Medicine and Health, Klinikum rechts der Isar, Research Group MITI \and Technical University of Munich, TUM School of Computation, Information and Technology, Artificial Intelligence in Healthcare and Medicine \and
Technical University of Munich, TUM School of Medicine and Health, Klinikum rechts der Isar, Department of Surgery \and
Imperial College London, Department of Computing
\\
\email{alexander.geiger@tum.de}}

\maketitle              
\begin{abstract}
High-resolution manometry (HRM) is the gold standard in diagnosing esophageal motility disorders. As HRM is typically conducted under short-term laboratory settings, intermittently occurring disorders are likely to be missed. Therefore, long-term (up to 24h) HRM (LTHRM) is used to gain detailed insights into the swallowing behavior. However, analyzing the extensive data from LTHRM is challenging and time consuming as medical experts have to analyze the data manually, which is slow and prone to errors. To address this challenge, we propose a Deep Learning based swallowing detection method to accurately identify swallowing events and secondary non-deglutitive-induced esophageal motility disorders in LTHRM data. We then proceed with clustering the identified swallows into distinct classes, which are analyzed by highly experienced clinicians to validate the different swallowing patterns. We evaluate our computational pipeline on a total of 25 LTHRMs, which were meticulously annotated by medical experts. By detecting more than 94\% of all relevant swallow events and providing all relevant clusters for a more reliable diagnostic process among experienced clinicians, we are able to demonstrate the effectiveness as well as positive clinical impact of our approach to make LTHRM feasible in clinical care.
\end{abstract}
\section{Introduction}
Benign esophageal diseases present significant health and socio-economic challenges, especially for aging populations. Dysphagia, characterized by difficulties in swallowing food, drinks or even saliva, becomes increasingly prevalent with age, posing challenges for patients and healthcare providers. While conditions such as gastroesophageal reflux disease have garnered considerable attention and research, disorders of esophageal motor function, such as dysphagia, remain less understood and frequently overlooked. Patients with sporadic esophageal disorders often experience severe symptoms, yet diagnosing these intermittent conditions can be immensely challenging.
High-resolution manometry (HRM) is the gold standard for diagnosing eso\-pha\-geal motility disorders. However, conventional HRM is typically conducted in controlled laboratory settings, following a standardized swallow protocol according to Chicago Classification (CC4.0) \cite{yadlapati:2021} within a limited timeframe. This approach is already proven to miss intermittent dysfunctions that manifest outside of the testing period \cite{keller:2020}. To address these limitations, we have extended HRM monitoring to a long-term setting (up to 24 hours) in a previous work \cite{jell:2016}. However, this expansion presents new challenges, as the increased data volume necessitates sophisticated methods for analysis and interpretation.
To address this challenge, in this paper we propose an automated swallow detection and clustering system, designed to streamline the analysis of longterm HRM (LTHRM) recordings, aiming to expedite the diagnostic process and alleviate the burden on healthcare professionals. The contributions of this work are:
\begin{itemize}
    \item We develop a novel Machine Learning (ML)-based procedure leveraging Convolutional Neural Networks (CNN), achieving a 94\% average Recall score for swallow detection, outperforming both a non-ML baseline and a commercial LTHRM evaluation tool.
    \item Our robust clustering approach categorizes detected swallows into distinct classes, reducing manual evaluation time for clinicians.
    \item We evaluate the performance on a total of 25 LTHRMs and provide an evaluation of the resulting clusters in terms of their clinical value by experienced clinicians.
\end{itemize}
\section{Related literature}
The analysis of HRM data includes a wide range of tasks such as automated sphincter motility analysis \cite{jungheim:2016,lee:2014} or probe position failure detection \cite{czako:2021}. Furthermore, the automatic analysis of swallows has been an active area of research, since the standardized HRM necessitates the manual evaluation of 10 to 21 swallow events per patient depending on the underlying pathology~\cite{yadlapati:2021}. It is widely acknowledged that there is faltering inter-rater reliability due to varying expertise levels in the evaluation of HRMs~\cite{fox:2015}. Consequently, various methodologies have been developed to aid and standardize this process, aiming for automatic categorization of these swallowing events according to the established CC4.0. Popa et al.~\cite{popa:2022} categorized 157 manometries, consisting of pre-labled swallows, into 10 distinct clusters utilizing a pretrained model. In their subsequent study, the classification was refined by combining multiple models aligning with a CC4.0 algorithm~\cite{surdea:2022}. Kou et al.~\cite{kou:2022} used a ML-based classification approach on standardized HRMs, which were manually labeled by medical experts.\par%
These studies have demonstrated robust methodologies for automated swallow classification achieving accuracies from 88\% to 97\%~\cite{kou:2022,kou:2022(2),popa:2022,surdea:2022}. Notably, these classifications are based on manually annotated individual swallow events that are separated by clearly defined intervals of at least 30 seconds. LTHRM goes beyond these limited laboratory studies leading to interlocking swallow events and 900 to 1500 swallows within 24 hours, thus making manual annotation impractical. \par%
In a previous work \cite{jell:2019}, we introduced an initial automated detection algorithm for identifying swallows during LTHRM assessment. This work is extended in this paper by using a Deep Learning based approach, enabling a more robust swallow event detection. Also, the detection is conducted in the whole sensor area and not just in the area of the previously manually defined upper esophageal sphincter. In addition, we thoroughly evaluate the resulting clusters in terms of their clinical value by experienced clinicians.
\section{Methodology}
\begin{figure}[t]
\centering
\includegraphics[width=\textwidth]{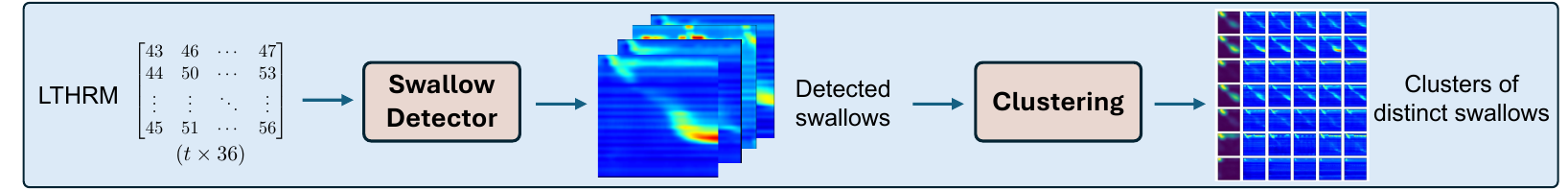}
\caption{Overview of our computational pipeline, consisting of a swallow detector and a subsequent clustering of the detected swallows.} 
\label{fig:overview}
\end{figure}

Our proposed computational pipeline employs a two step approach. The first step involves automated detection of all swallowing events in the manometric data. Similar swallows are subsequently clustered into groups, resulting in the presentation of only a few representative images of each swallow cluster to medical experts for further diagnosis. The overall approach is depicted in Figure \ref{fig:overview}.
\subsection{Data set}
The data set consists of 25 LTHRMs of patients with suspected intermittently occurring motility disorders of the esophagus, collected at *** University Hospital. The patient data was collected and used with ethical approval and informed consent. All patients underwent endoscopy prior to HRM to rule out malignancy and other structural causes of dysphagia. After a fasting period of at least 6 hours, the manometry catheter was placed transnasally and a standardized examination was performed according to CC4.0 protocol \cite{yadlapati:2021}. In absence of functional reasons for dysphagia, the patients were introduced to the specifics of the extended LTHRM examination. They were advised to keep a recording on their meals, body position, symptoms and maintain their daily routine as much as possible to obtain representative and traceable measurements.\par%
The HRM catheter contains 36 circumferential pressure sensors located 1 cm apart. Manometric values are collected at a sampling rate of 50 Hz, resulting in a manometry matrix denoted $\mathbf{M} \in \mathbb{R}^{36 \times t}$, where $t$ is the number of measurements. The values are smoothed by using a moving average window across the $t$ dimension to get $\mathbf{\hat{M}}$, where $\mathbf{\hat{M}}_{i,j} = \frac{1}{w} \sum_{k=j}^{j+w-1} \mathbf{M}_{i,k}$, with $i = 1 \dots 36; j = 1 \dots t-w+1$, and $w=30$ in our experiments. The values in $\mathbf{\hat{M}}$ are then clipped between -200 and 300 mmHg to remove any extreme values and scaled to a standardized range between 0 and 255. After collection, each data set was evaluated by experienced clinicians and each swallow event start was carefully labeled, resulting in a total of over 25,000 labeled swallow events.
\subsection{Swallow detection}
The initial step involves the automated detection of swallow events in LTHRM. For this purpose, we implement and compare several approaches that are described below.

\subsubsection{Non-ML baseline: Threshold-based approach}

As the pressure along the 36 sensors behaves characteristically during swallowing (e.g. pressure increase in the pharynx/ esophagus), we use a threshold-based approach to find peaks in the pressure values to identify swallow events. This serves as a non-ML baseline. The method takes the complete preprocessed manometry $\mathbf{\hat{M}}$ as an input. This matrix is then converted to a binary matrix $\mathbf{M_b}$ and a moving sum is applied to highlight regions with a series of consecutive 1s before summing across all sensors. Formally,
\begin{align*}
\mathbf{M_b}_{i,j} =
\begin{cases} 1, & \text{if } \mathbf{\hat{M}}_{i,j} > \text{80} \\0, & \text{otherwise}
\end{cases};
\qquad 
\mathbf{M_s}_{i,j} = \sum_{k=i}^{i+w-1} \mathbf{M_b}_{k,j};
\qquad \mathbf{r}_j = \sum_{k=1}^{36} \mathbf{M_s}_{k,j}
\end{align*}

where $i,j$ are the same as before and $w=20$. The resulting vector $\mathbf{r} = [\mathbf{r}_1, \ldots, \mathbf{r}_t] \in \mathbb{R}^{1 \times t}$ is smoothed again to get the final vector $\mathbf{\hat{r}}$, such that $\mathbf{\hat{r}}_j = \frac{1}{w} \sum_{k=j}^{j+w-1} \mathbf{r}_{k}$ for  $ j = 1 \dots t-w+1$, and $w=100$. In this vector, a peak finding algorithm detects all peaks as swallows if their value is $>20$ and their distance is $>200$, which proved to be reasonable values in our experiments.

\subsubsection{ML-based approach}

We use a supervised learning approach to classify an input manometry sequence $\mathbf{I} \in \mathbb{R}^{36 \times 500}$ into one of two classes, where 1 resembles a swallow sequence and 0 a non-swallow sequence. We created a training set by automatically iterating over the annotated manometries $\mathbf{\hat{M}}$ and storing a window of length 500 beginning from each annotated swallow start. This results in a swallow tensor $\mathbf{S} \in \mathbb{R}^{s \times 36 \times 500}$, with $s$ being the number of swallows in the data set $\mathbf{\hat{M}}$. For the non-swallow events $\mathbf{N}$, we sample windows of the same dimension from locations in $\mathbf{\hat{M}}$ that are located between annotated swallows. In order to get the required input dimensions for the specific models, each sequence $\mathbf{I}_i \in \{\mathbf{S},\mathbf{N}\}$ is re-scaled such that $\mathbf{I}_i^{36 \times 500} \rightarrow \mathbf{I}_i^{224 \times 224}$.
We implement several CNNs to compare their performance in our specified task, namely GoogLeNet~\cite{szegedy:2015}, MobileNet~\cite{howard:2019}, EfficientNet~\cite{tan:2019}, and RegNet~\cite{radosavovic:2020}. The models were pre-trained on the ImageNet data set \cite{deng:2009}. The dimension of the fully connected layer of each model was adapted to match our two binary output classes.
The models were trained using the stochastic gradient descent optimizer with a learning rate of 3e-3 and a batch size of 128 for 20 epochs. We report the test results of the models that performed best on the validation set for each fold. The models were implemented in PyTorch and trained on a NVIDIA RTX A6000.\par
\begin{figure}[t]
\centering
\includegraphics[width=0.9\textwidth]{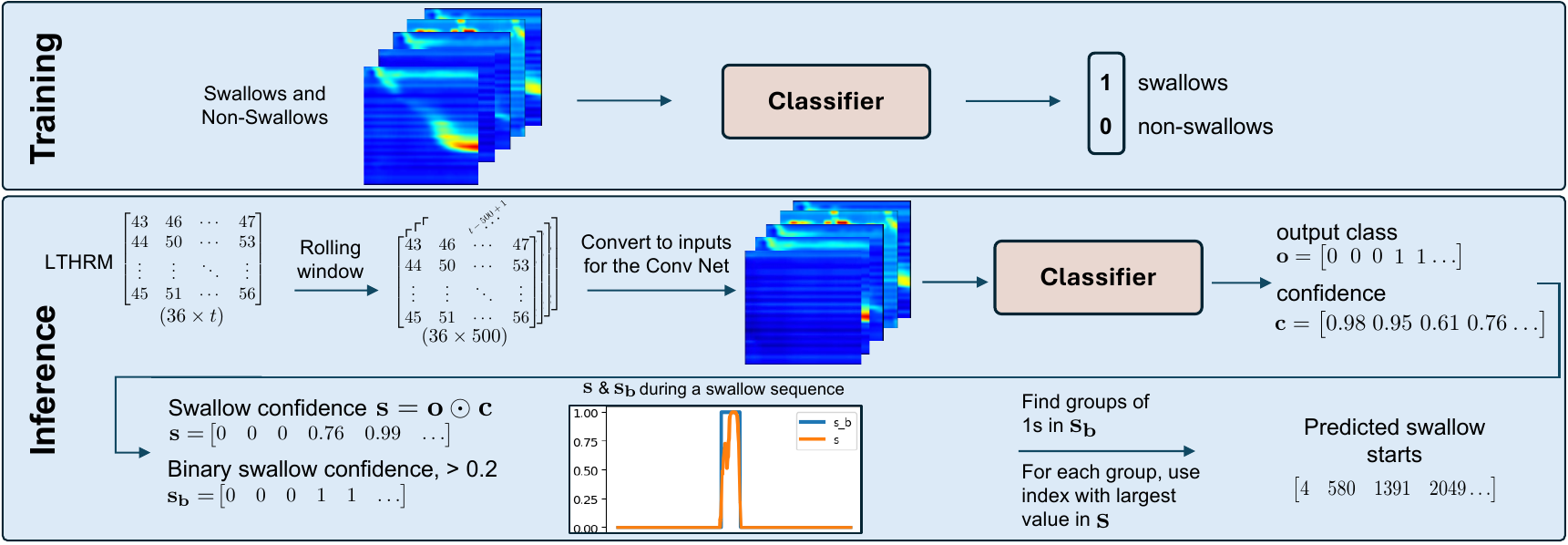}
\caption{The training and inference procedure of our swallow detection method.} 
\label{fig:swallow_detection}
\end{figure}

%
For the inference part, we developed a detection pipeline by applying a rolling window of length 500 over the test manometry $\mathbf{\hat{M}}$, resulting in $t-500+1$ inference windows, denoted as $\mathbf{I}_i$ for $i=0,\ldots,t-500+1$, which are passed to the classifier model, such that for each $\mathbf{I}_i$ the output class and the corresponding confidence is computed. The resulting two vectors are the binary output class vector $\mathbf{o}$ and confidence vector $\mathbf{c}$, with $\mathbf{o},\mathbf{c} \in \mathbb{R}^{1 \times (t-500 + 1)}$. We then use the element-wise product $\mathbf{s}=\mathbf{o} \odot \mathbf{c}$ to obtain only the confidences of class 1 outputs. Afterwards, a moving averaging window is applied to smooth the values and convert the resulting vector $\mathbf{\hat{s}}$ to a binary vector $\mathbf{s_b}$, formally
\begin{align*}
\mathbf{\hat{s}} = [\hat{s}_1,\ldots,\hat{s}_{j}], \text{with } \mathbf{\hat{s}}_j = \frac{1}{w} \sum_{k=j}^{j+w-1} \mathbf{s}_{k};
\qquad 
\mathbf{s_b}_{i} =
\begin{cases} 1, & \text{if } \mathbf{\hat{s}}_{i} > \text{0.2} \\0, & \text{otherwise}
\end{cases}
\end{align*}
with $w=20$. All groups of consecutive 1s in $\mathbf{s_b}$ are considered a detected swallowing sequence. Using the maximum value in $\mathbf{s}$ indicates the predicted start of a swallow sequence. The training and inference pipeline is depicted in Figure \ref{fig:swallow_detection}.
\subsection{Clustering of similar swallows}
\begin{figure}[t]
\centering
\includegraphics[width=\textwidth]{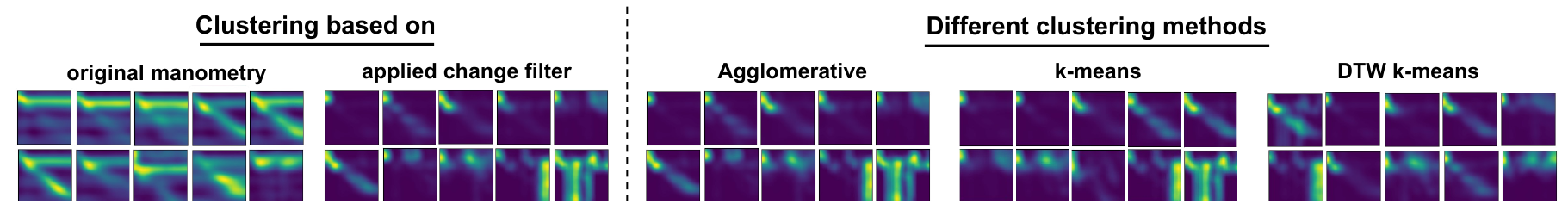}
\caption{Comparison of different clustering options. Left: Difference between using the pure manometry values and applying the change filter. Right: Comparison if different clustering methods - agglomerative clustering and k-means achieve similarly distinctive clusters, while DTW based k-means results in slightly less distinctive clusters.}
\label{fig:clustering_comparison}
\end{figure}

To cluster the detected swallows, we compared multiple clustering methods, including standard k-means, agglomerative clustering, and Dynamic Time Warping (DTW) based k-means. In our evaluation, k-means and agglomerative clustering tended to result in very similar clusters, while the DTW based k-means method took longer and produced slightly more homogeneous clusters, therefore favoring the first two methods. For the remainder of the evaluation, we decided to focus on the agglomerative clustering. Additionally, we compare to cluster the plain swallow images as well as applying a change filter to the images beforehand, which specifically highlights pressure changes during a swallow. Based on a qualitative evaluation, we observed the best results using a change filter kernel $k$ of the form $ k = [-1, 0, ...,0,1] \in \mathbb{R}^{1 \times 10}$, which is convoluted across manometry $\mathbf{I}_i$ for each swallow $i$. In essence, this filter aims to detect changes in the sensor values within a time frame of 10 measurements. The resulting change matrices $\mathbf{C}_i$ are squared and then re-scaled such that $\mathbf{C}_i^{36 \times 500} \rightarrow \mathbf{C}_i^{50 \times 50}$, before applying a Gaussian filter for a final smoothing. Figure \ref{fig:clustering_comparison} shows a comparison of the different clustering methods, as well as the effect of using the pure manometry values compared to applying the specified change filter. 

Prior to clustering, the matrices are flattened and a Principal Component Analysis (PCA) is performed to reduce the dimensions such that $\mathbf{C}_i^{50 \times 50} \rightarrow \mathbf{c}_i^{2500 \times 1} \rightarrow \text{PCA} \rightarrow \mathbf{c}_i^{30 \times 1}$. This results in the final vectors $\mathbf{c}_i$ for each detected swallow $i$, which are then passed to the clustering algorithm. The initial number of clusters is determined by running the clustering multiple times with different cluster numbers (from 4 to 10 clusters) and selecting the number with the lowest mean intra-cluster distance.\par%

We consider all clusters containing 15\% or more of all samples to be the main categories of swallows. As clinicians are mainly interested in special cases when looking for intermittent occurring motility disorders, the remaining samples are clustered a second time. A predefined cluster number of 10 is utilized, effectively creating distinct separations among clusters without resulting in an excessive number of similar groupings.

\section{Results}

\subsection{Swallow detection}
A comparative analysis is conducted between the introduced detection methods and the only commercially available LTHRM evaluation software (ViMeDat v5.1.6.0, Standard Instruments, Germany). The evaluation aims to gauge the efficacy of the proposed method within the medical context. To measure the performance of the methods we deploy three different metrics suitable for classification tasks, i.e. precision, recall, and F1-score. Our ML-based approaches and ViMeDat aim to detect the swallowing start. Therefore, we count a correct detection if a predicted swallow start is in the range $[y - \frac{1}{2}d,\ldots, y+\frac{1}{2}d]$ with $y$ being the true swallow start and $d=400$. In contrast, the baseline approach aims to detect swallows during the swallow event, therefore we count a correct detection if a predicted swallow is in the range $[y,\ldots, y+d]$. In the supplementary material, we provide an overview of the distributions of errors around the correct starts for the different methods as well as additional results for different values of $d$.
We compare the different methods using a 5-fold cross validation, where each fold contains the LTHRM of 5 patients. It can be observed that the ML-based algorithms achieve the highest scores (see Table \ref{tab1}), while MobileNet scores highest in all metrics with a precision of 86.1\%, recall of 94.1\% and F1-Score of 89.6\%. The ML-based algorithms outperform both the non-ML baseline as well as the commercially available ViMeDat tool. It is noteworthy that the detection within this tool is tied to the initial manual annotation of anatomical features (i.e. localisation of sphincters) by medical experts, while our proposed ML-based approach is fully automated. 

\begin{table}[t]
\begin{center}
\caption{Comparison of the different detection methods. We report the average metrics using a 5-fold cross validation along with their respective standard deviation ($\pm$).}
\label{tab1}
\begin{tabular}{l @{\hskip 0.35in} c  @{\hskip 0.3in} c  @{\hskip 0.3in} c }
\hline
Method & Precision (\%) & Recall (\%) & F1-Score (\%) \\
\hline
Non-ML Baseline & $29.58 \pm 4.50$ & $76.70 \pm 9.68$ & $38.70 \pm 7.70$ \\
ViMeDat & $85.73 \pm 4.49$ & $54.18 \pm 15.86$ & $61.59 \pm 16.19$ \\
GoogLeNet & $83.48 \pm 4.39$ & $94.05 \pm 2.16$ & $88.00 \pm 3.23$ \\
MobileNet & $\mathbf{86.13 \pm 2.98}$ & $\mathbf{94.07 \pm 2.46}$ & $\mathbf{89.57 \pm 2.59}$ \\
EfficientNet & $83.86 \pm 5.55$ & $94.01 \pm 1.80$ & $88.23 \pm 4.06$ \\
RegNet & $80.94 \pm 4.59$ & $91.06 \pm 3.96$ & $85.27 \pm 4.25$ \\
\hline
\end{tabular}
\end{center}
\end{table}

\subsection{Clustering}
\begin{figure}[t]
\centering
\includegraphics[width=0.95\textwidth]{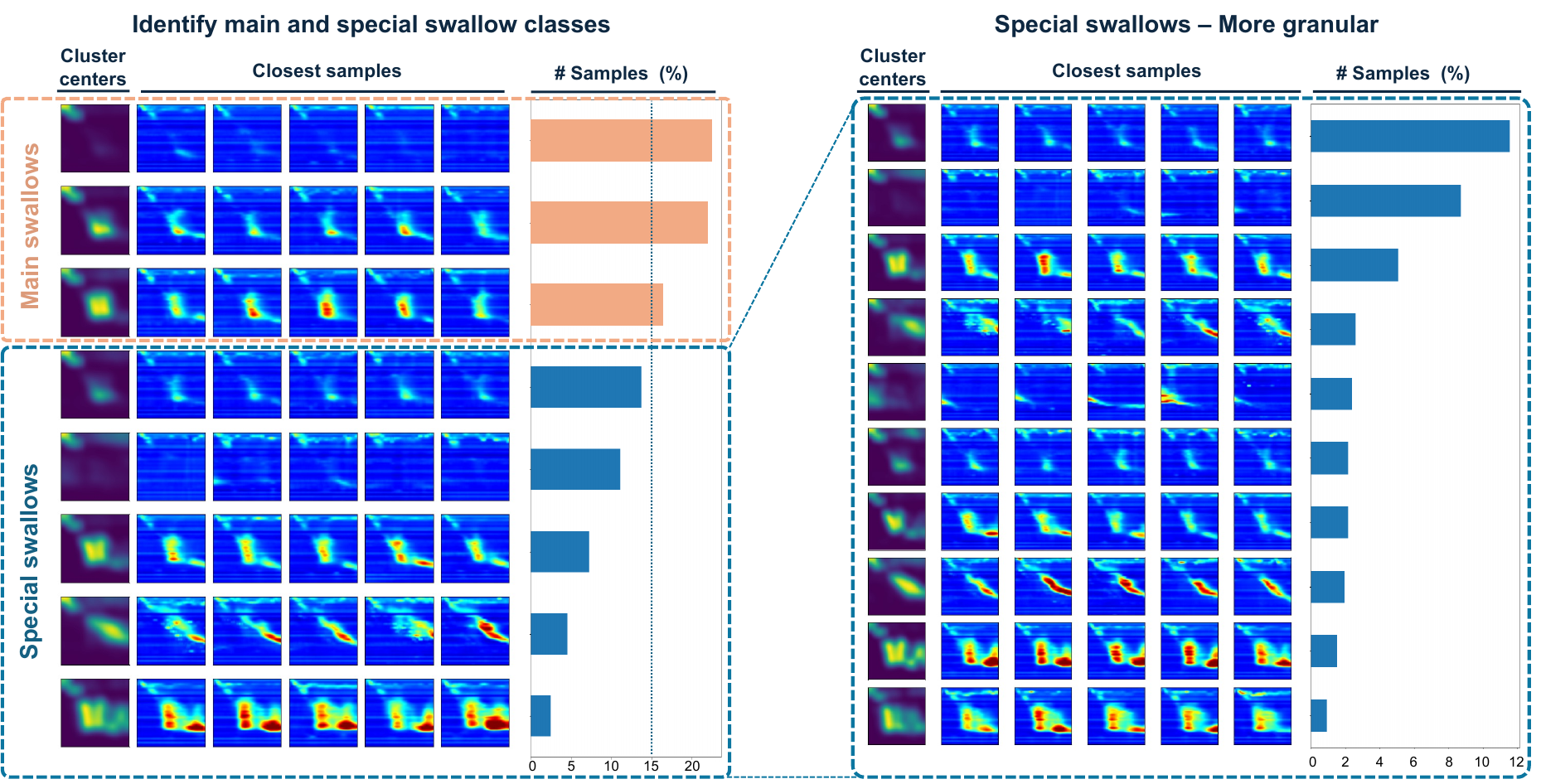}
\caption{Clustering of all detected swallows for a single patient. Special classes with an occurrence $<15\%$ are clustered more granularly on the right side. 
}
\label{fig:clustering}
\end{figure}

The clustering of the extracted swallows was evaluated using a randomized evaluation study with 5 healthcare professionals with in-depth knowledge of HRM. Each of these professionals received two versions of 5 randomly selected LTHRM cases: (1) a complete LTHRM presented in the commercially available ViMeDat software; (2) the clustered swallows (see Figure \ref{fig:clustering}). We evaluated time savings as well as subjective confidence and reliability in diagnostic findings.\par%
All professionals agreed on a significant time reduction due to automated detection and clustering, although screening the LTHRM for potentially undetected swallow events remained necessary. However, this fact was not considered critical by the medical professionals. We used the Fleiss' kappa coefficient $\kappa$ \cite{landis:1977} to measure the inter-rater reliability for the diagnostic findings by the medical experts for both versions. Among all conventionally analysed LTHRMs by means of version (1), inter-rater reliability for detecting all pathologies achieved $\kappa = 0.53$, whereas diagnostic findings by means of the clustered swallows (2) achieved $\kappa=0.73$. This improvement in inter-rater reliability through the use of automated detection and clustering in the analysis of LTHRMs indicates a more consistent and reliable diagnostic process among medical professionals. While the responsibility for ML-based diagnosis remains a critical aspect that requires careful monitoring and clear guidelines to ensure ethical and effective use \cite{ho:2019,neri:2020}, our results show the potential for ML-enhanced patient care through the integration of such technologies.
\section{Conclusion}
In this work we propose a novel swallow detection and clustering method in LTHRMs, allowing for automated assessment of all relevant swallow events of patients with suspected intermittently occurring motility disorders of the esophagus. Our ML-based swalllow detection algorithm achieves a recall of 94.07\%, surpassing the performance of existing commercial software solutions. The subsequent clustering algorithm demonstrates its effectiveness in providing a more reliable diagnostic process among medical professionals by increasing the Fleiss' kappa coefficient by 0.2 compared to conventional diagnostic methods. Consequently, we are able to demonstrate the effectiveness as well as clinical positive impact of our approach to make LTHRM feasible in clinical care.\par%
In a future step, we plan to improve the proposed algorithm on a broader data set to integrate not only clustering but also automatic classification of swallows into predefined classes. This development promises to further streamline the diagnostic process, enabling more precise and efficient identification of swallow characteristics. Ultimately, this could lead to significant advancements in the treatment of swallowing disorders, by providing healthcare professionals with a more nuanced understanding of swallow patterns.

%
%
%
\bibliographystyle{splncs04}
\bibliography{bibliography}

\newpage
\appendix

\section{Distances of predicted swallows to true swallows}

\begin{figure}[H]
\centering
\includegraphics[width=\textwidth]{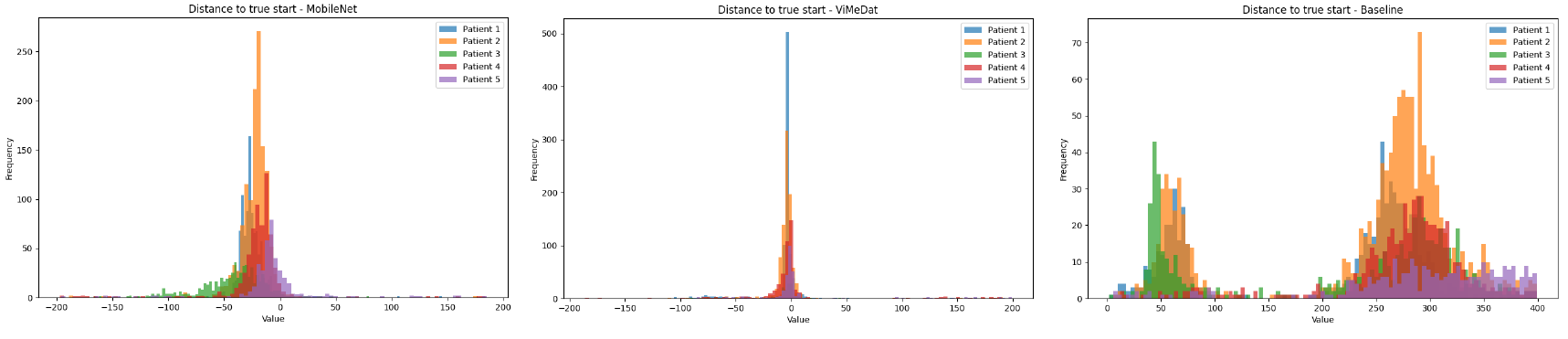}
\caption{The distance between the detected swallow and the correct swallow start for 5 patients. It can be observed, that the distances of the MobileNet outputs are centered around -35, indicating that our approach is typically predicting the start of a swallow slightly earlier compared to the labelled start. The distances of the ViMeDat outputs are centered around 0, indicating that the software is typically predicting the swallow exactly at the true swallow start. The baseline, as it is looking at high pressure events which mostly occur during the swallow, is predicting the swallow event after the swallow start. It can be observed that most of the times a high pressure event is either occurring rather soon after the labeled swallow start (50 measurements or 1 second), or after around 270 measurements or 5-6 seconds after the swallow start.
}
\label{fig:distances}
\end{figure}

\newpage

\section{Different values of allowed range $d$}
\begin{table}[H]
\begin{center}
\caption{Comparison of the different detection methods, using $d=100$. For the MobileNet and ViMeDat approach, this means that a correct swallow detection is counted if the predicted swallow start is in a distance of at most $\pm50$ measurements from the true swallow start. As the calculation of the Baseline is not focusing on detecting the start of a swallow, but rather the actual swallow event, reducing the allowed distance is not applicable here. We report the average metrics over a 5-fold cross validation along with their respective standard deviation ($\pm$).}\label{tab2}
\begin{tabular}{l @{\hskip 0.35in} c  @{\hskip 0.3in} c  @{\hskip 0.3in} c }
\hline
Method & Precision (\%) & Recall (\%) & F1-Score (\%) \\
\hline
Non-ML Baseline & n/a & n/a & n/a \\
ViMeDat & $\mathbf{79.62} \pm 5.12$ & $50.28 \pm 14.05$ & $57.23 \pm 14.36$ \\
MobileNet & $77.50 \pm 7.63$ & $\mathbf{84.68 \pm 6.49}$ & $\mathbf{80.62 \pm 7.18}$ \\
\hline
\end{tabular}
\end{center}
\end{table}

\begin{table}[H]
\begin{center}
\caption{Comparison of the different detection methods, using $d=800$. For the MobileNet and ViMeDat approach, this means that a correct swallow detection is counted if the predicted swallow start is in a distance of at most $\pm400$ measurements from the true swallow start. As the calculation of the Baseline is not focusing on detecting the start of a swallow, but rather the actual swallow event, a correct swallow detection in this case is counted if the predicted swallow event is in range $[y-200,\ldots, y+600]$ with $y$ being the true swallow start.  We report the average metrics over a 5-fold cross validation along with their respective standard deviation ($\pm$).}\label{tab3}
\begin{tabular}{l @{\hskip 0.35in} c  @{\hskip 0.3in} c  @{\hskip 0.3in} c }
\hline
Method & Precision (\%) & Recall (\%) & F1-Score (\%) \\
\hline
Non-ML Baseline & $36.37 \pm 4.76$ & $86.32 \pm 9.29$ & $46.67 \pm 8.56$ \\
ViMeDat & $88.47 \pm 3.86$ & $58.46 \pm 16.95$ & $65.37 \pm 16.66$ \\
MobileNet & $\mathbf{89.04 \pm 1.92}$ & $\mathbf{97.65 \pm 1.19}$ & $\mathbf{92.82 \pm 1.36}$ \\
\hline
\end{tabular}
\end{center}
\end{table}

\newpage

\section{Example of most distant samples for each cluster}

\begin{figure}[!h]
\centering
\includegraphics[width=0.9\textwidth]{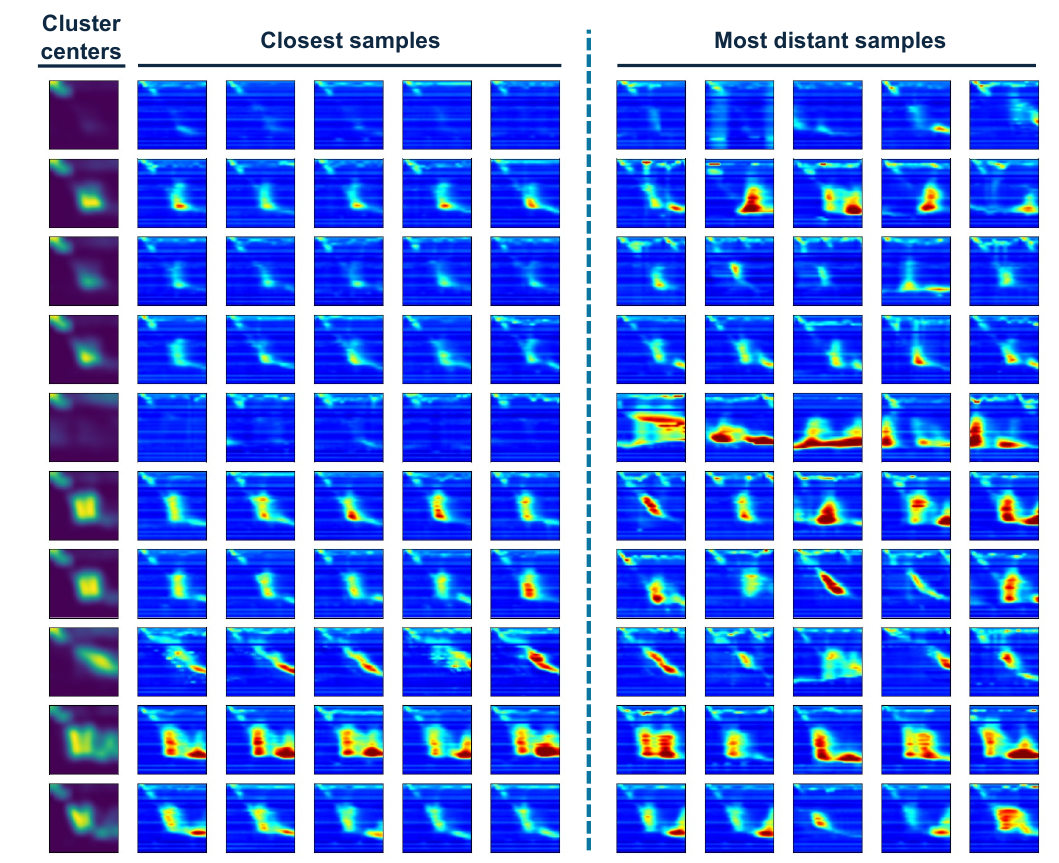}
\caption{The closest swallows to their respective cluster center as well as the most distant samples. It can be observed that for the majority of clusters the most distant samples still objectively belong to this cluster, indicating that the produced cluster centers are indeed a representative and sufficient description of the different swallow types present in the LTHRM.
}
\label{fig:distantsamples}
\end{figure}

\end{document}